\documentclass{article}
\usepackage{ijcai24}
\usepackage{times}
\usepackage{soul}
\usepackage{url}
\usepackage[hidelinks]{hyperref}
\usepackage[utf8]{inputenc}
\usepackage[small]{caption}
\usepackage{graphicx}
\usepackage{amsmath}
\usepackage{amssymb}
\usepackage{amsthm}
\usepackage{booktabs}
\usepackage{algorithm}
\usepackage[switch]{lineno}
\usepackage{algpseudocode}
\usepackage{subfig}

\title{Effect of Adaptation Rate and Cost Display in a Human-AI Interaction Game}

\author{
Jason T. Isa$^1$
\and
Bohan Wu$^1$\and
Qirui Wang$^1$\and
Yilin Zhang$^1$\and
Samuel A. Burden$^1$\and
Lillian J. Ratliff$^1$\And
Benjamin J. Chasnov$^1$\\
\affiliations
$^1$University of Washington, Seattle\\
\emails
\{jisa, bohanw, qw43, yilinz24, sburden, ratliffl, bchasnov\}@uw.edu,
}

\begin{document}

\algnewcommand\algorithmicswitch{\textbf{switch}}
\algnewcommand\algorithmiccase{\textbf{case}}
\algnewcommand\algorithmicassert{\texttt{assert}}
\algnewcommand\Assert[1]{\State \algorithmicassert(#1)}%
\algdef{SE}[SWITCH]{Switch}{EndSwitch}[1]{\algorithmicswitch\ #1\ \algorithmicdo}{\algorithmicend\ \algorithmicswitch}%
\algdef{SE}[CASE]{Case}{EndCase}[1]{\algorithmiccase\ #1}{\algorithmicend\ \algorithmiccase}%
\algtext*{EndSwitch}%
\algtext*{EndCase}%

\maketitle

\begin{abstract}
    As interactions between humans and AI become more prevalent, it is critical to have better predictors of human behavior in these interactions. We investigated how changes in the AI's adaptive algorithm impact behavior predictions in two-player continuous games. In our experiments, the AI adapted its actions using a gradient descent algorithm under different adaptation rates while human participants were provided cost feedback. The cost feedback was provided by one of two types of visual displays: (a) cost at the current joint action vector, or (b) cost in a local neighborhood of the current joint action vector. Our results demonstrate that AI adaptation rate can significantly affect human behavior, having the ability to shift the outcome between two game theoretic equilibrium. We observed that slow adaptation rates shift the outcome towards the Nash equilibrium, while fast rates shift the outcome towards the human-led Stackelberg equilibrium. The addition of localized cost information had the effect of shifting outcomes towards Nash, compared to the outcomes from cost information at only the current joint action vector. Future work will investigate other effects that influence the convergence of gradient descent games.
\end{abstract}

\section{Introduction}

\begin{figure}[t]
    \center
    \includegraphics[width=0.84\linewidth]{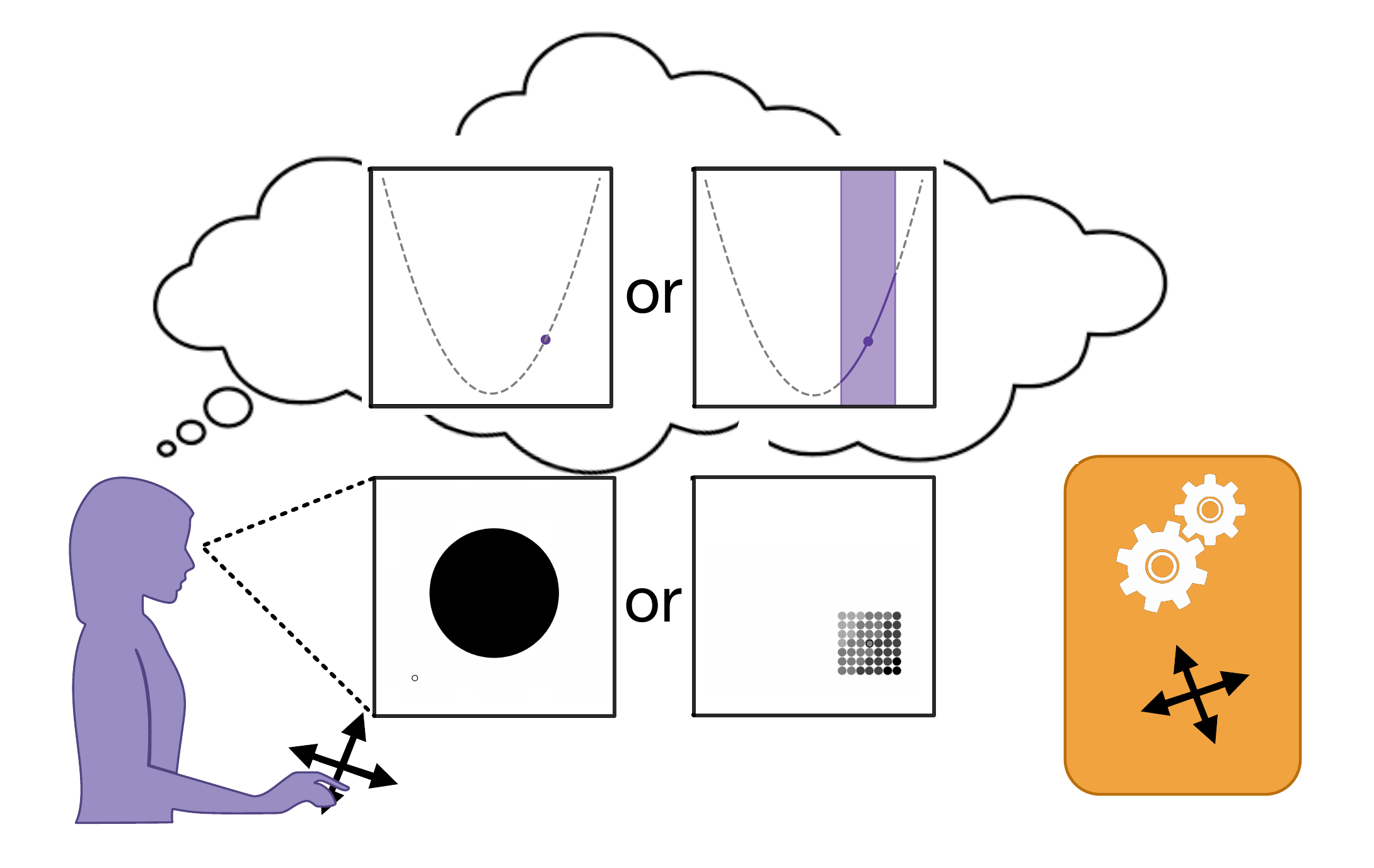}
    \includegraphics[width=0.77\linewidth]{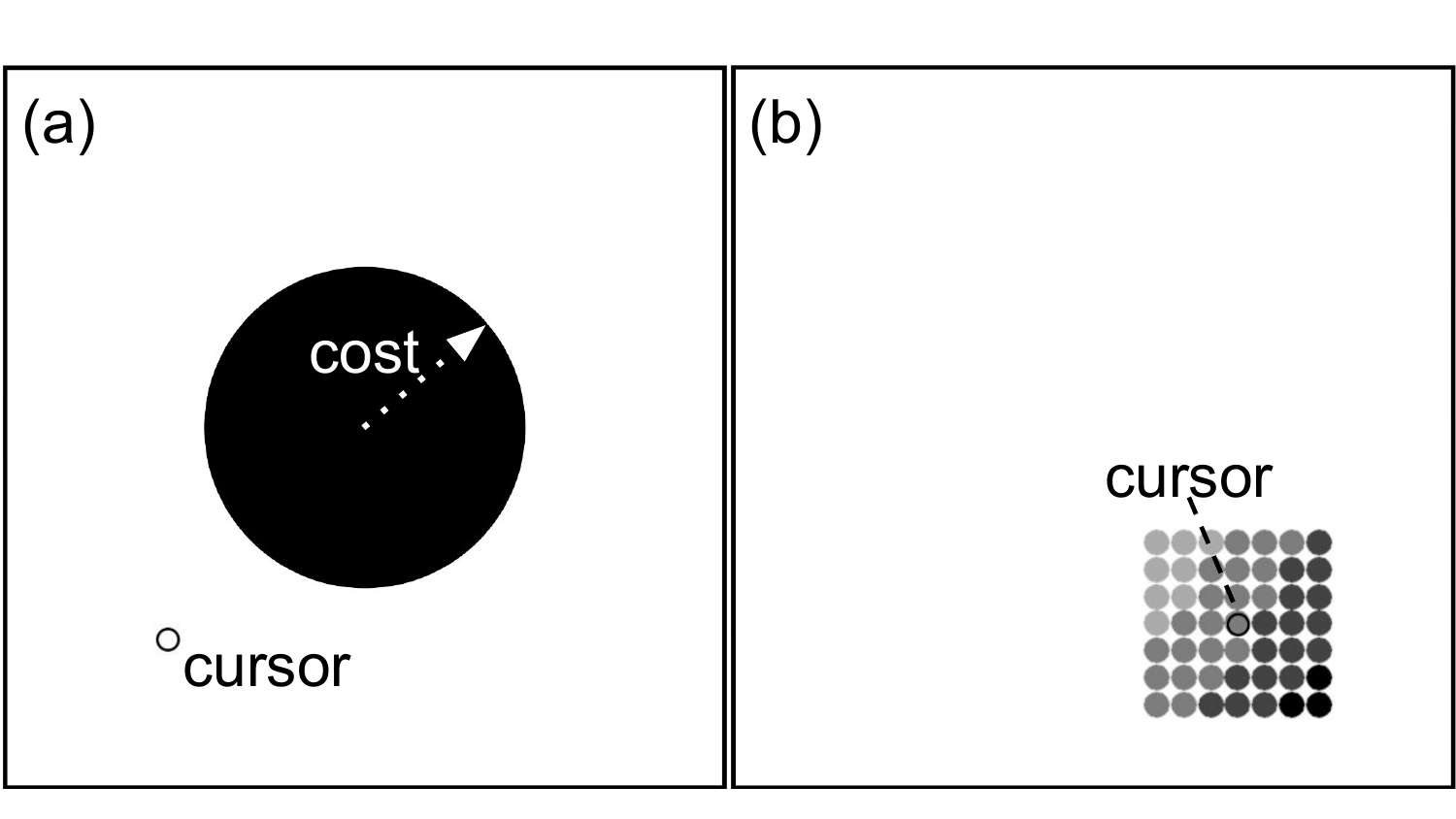}
    \caption{Human participants played a continuous game with an adaptive AI where they were shown one of two display interfaces: (a) current human cost (Experiment 1 annotated screenshot), similar to seeing a single point on their cost landscape, and (b) localized heat map (Experiment 2 annotated screenshot), similar to seeing a localized subsection of their perceived cost landscape.}
    \label{fig:setup}
\end{figure}

Interactions between humans and AI are increasing in our daily lives~\cite{6945523,{doi:10.1080/001401300409044},book,cannan2011human}. To ensure that AI can be designed appropriately~\cite{boy2017handbook}, it is important to have better predictors of human behavior in interactive tasks. In this paper, we leverage game-theoretic equilibria based on Nash~\cite{nash1950equilibrium} and Stackelberg~\cite{von2010market} equilibria to predict the outcome of a human-AI game. Many human-AI interactions can be modeled as a game~\cite{von1947theory}. In particular, simultaneous play games can be modeled as two players performing gradient descent over their own actions, while Stackelberg game can be modeled as a hierarchical game between a leader and a follower, where the leader performs gradient descent assuming the follower plays its best response. In this work, we show how the adaptation rate of the AI's gradient descent algorithm affects the joint action vector outcome of the game, modeled by these two game-theoretic equilibria. We then show how these predicted outcomes can change when the human is presented with different feedback information.

This paper considers a two-player general-sum game between a human and an AI. The game is played repeated over two different user cost displays. In one display, human participants explored their input space, similar to seeing only a single point on their cost landscape, Figure~\ref{fig:setup}a. While in the other display the human had additional information about their perceived localized cost landscape, from which one could potentially estimate cost differences or gradient information, Figure~\ref{fig:setup}b. Within the game, participants were instructed to optimize an objective while the AI adapts behind-the-scene. This system serves as a test-bed for studying game-theoretic learning dynamics and behavior outcomes in human-AI interaction that can be generalized to various domains involving adaptive machine learning algorithms that learns from human data.

There are many prior works on human-AI interaction~\cite{march2021strategic,rosenfeld2018strategic}. Some research on user interfaces require users to choose discrete actions~\cite{10.2307/2234532,crandall2018cooperating}. However, in our work, the user repeatedly chooses a continuous action, for example, like this related research on human-AI negotiation~\cite{doi:10.1080/10864415.2020.1767428}. Other research are more similar to ours, such as~\cite{nikolaidis2017game,li2019differential,7548305}, where players play dynamic games over continuous action spaces. Studying continuous action spaces offers advantages due to realistic modeling of real-world scenarios. Many real-world multi-agent settings involve continuous actions, such as in robotics, control systems, and learning algorithms. Continuous actions can also be generalized to discrete actions by using continuous variables to represent discrete actions by smooth approximations or mixed strategies. 

While many studies in this field have studied models based on cooperative or zero-sum games~\cite{lai2021science,10.1145/3512930}, our research adopts a general-sum framework, offering a more realistic representation of feedback loops in human-AI interactions~\cite{bacsar1998dynamic}. Moreover, unlike existing works that focus on how information about the AI's outputs is displayed to the human user~\cite{bondi2022role,10.1145/3449287}, our study modifies the information that humans receive about their own objectives. There are also many prior works on studying equilibrium outcomes in games and learning dynamics~\cite{march2021strategic}. Furthermore, behavioral studies~\cite{Camerer2010} and empirical studies~\cite{chugunova2022we} have shown that human adaptation in some settings can be modeled and empirically measured. In our study, we aim to empirically validate the game-theoretic model through crowd-sourced experiments. This research could be used to motivate possible design choices in human-AI systems.

We showed, through human subjects experiments and numerical simulation, that the adaptation rate had a robust effect of shifting the behavior outcome of human actions from the Nash equilibrium to the human-led Stackelberg equilibrium. As the AI's adaptation rate increased, human participants' cost decreased, leading the participant to a more human beneficial outcome while the AI's cost stayed relatively constant. 

In what follows, we will first provide a formal definition of the games we study and the relevant game-theoretic equilibrium concepts. Following, we describe the quadratic cost model and cost feedback display setup for the experiments. We include algorithm definitions for the human experiments and simulations. Our results highlight the effect of varying the AI's adaptation rate while participants used different feedback displays. We find that both the rate and cost display had significant effects on the human action vector outcome of the game.

\section{Preliminaries}
We now formalize the games we studied and present equilibrium concepts accompanied by sufficient condition characterizations.

Consider a non-cooperative game between two agents where the Human agent $H$ plays the AI agent $M$. The Human agent has cost $c_H:\mc H \times \mc M \to\R $ and the AI agent has cost $c_M:\mc H\times \mc M \to \R$  where $\mc H \subseteq \R^{d_H}$ and $\mc M \subseteq \R^{d_M}$ denoting the action spaces of the respective agents such that $d_H,d_M$ denote the number of action inputs the respective agents have. We assume throughout that each $c_H, c_M$ is sufficiently smooth: $c_H,c_M \in C^q(X,R)$ for some $q\geq 2$. In words, we consider the class of two-player smooth games on continuous, constrained actions spaces. 

A joint action $(h^\NE, m^\NE) \in \mc H \times \mc M$ is a \emph{Nash equilibrium} if $c_H(h^\NE, m^\NE) \leq c_H(h, m^\NE)$, $\forall h\in \mc H$ and $c_M(h^\NE, m^\NE) \leq c_M(h^\NE, m)$, $\forall m\in \mc M$. A joint action  $(h^\SE, m^\SE) \in \mc H \times \mc M$ is a \emph{Stackelberg equilibrium} with the Human agent as the leader if $\forall h\in \mc H$, $\sup_{m \in M(h)} c_H(h^\SE,m) \leq \sup_{m\in M(h) }c_H(h,m)$ where $M(h)=\{m' \in \mc M \mid c_M(h,m') \leq c_M(h,m) \forall m \in \mc M \}$. Refer to~\cite{fiez2020implicit} for the definition of Nash and Stackelberg equilibria. 

In a simultaneous play game ~\cite{bacsar1998dynamic}, 
the agents aim to solve optimization problems,
\begin{align*}
h^\NE &= \arg\min_{h \in\mc H} c_H(h, m^\NE),\\
m^\NE &= \arg\min_{m\in\mc M} c_M(h^\NE, m)
\end{align*}
where $(h^\NE, m^\NE) \in \mc H \times \mc M$
is the Nash equilibrium. In this setting, agents are modeled as making decisions simultaneously, in contrast to the sequential nature of Stackelberg games.

In the formulation of a Stackelberg game~\cite{bacsar1998dynamic},
the leader makes the first move, then the follower responds. In the modern machine learning setting, recent work has shown that Stackelberg equilibria arise for gradient-based updates~\cite{fiez2020implicit}.
In a Stackelberg game, the leader and follower aim to solve
the following optimization problems:
\begin{align*}
h^\SE &= \arg\min_{h\in\mc H} \{ c_H(h,m^\SE) \mid m^\SE=\arg\min_{m} c_M(h, m)\},\\
m^\SE &= \arg\min_{m\in\mc M} \{ c_M(h^\SE,m)\}.
\end{align*}
The designation of ‘leader’ and ‘follower’ indicates
the order of play between the agents, meaning the leader
plays first and the follower second.
In our experiments, the leader that emerges is the Human agent, and the AI agent is the follower.

We remark that Nash equilibria exist for convex costs on compact and convex action spaces ~\cite{8344bf6b-b18f-3bf0-88f4-c6a0b490d844} while Stackelberg equilibria exist on compact action spaces~\cite[Thm 4.3, Thm 4.8 \& Section 4.9]{bacsar1998dynamic}. Predicated on existence, equilibria can be characterized in terms of sufficient conditions on agent costs.  We denote  $\frac{\partial c_H}{\partial h}$ as the derivative of $c_H$ with respect to $h$ and $\frac{\partial c_M}{\partial m}$ as the derivative of $c_M$ with respect to $m$. The following gives sufficient conditions for a Nash equilibrium.

[Sufficient conditions for differential Nash equilibrium~\cite{7028565}]
The joint strategy $(h^\NE, m^\NE)\in\mc H \times \mc M$
is a \emph{differential Nash equilibrium} if
\begin{align}
\frac{\partial c_H}{\partial h}(h^\NE,m^\NE)&=0,\ 
&\frac{\partial c_M}{\partial m}(h^\NE,m^\NE)=0,\\
\frac{\partial^2 c_H}{\partial h^2}(h^\NE,m^\NE)&\succ 0,\ 
&\frac{\partial^2 c_M}{\partial m^2}(h^\NE,m^\NE)\succ 0.
\end{align}

Analogous sufficient conditions can be stated to characterize a Stackelberg equilibrium. Towards this end, 
let the best response of the AI agent be  $\BR_M(h) = \arg\min_{m} c_M(h,m)$. 
We denote $\frac{\partial c_H}{\partial h}(\cdot ,\BR_M(\cdot ))$ as the total derivative of $c_H$.

[Sufficient conditions for differential Stackelberg equilibrium~\cite{fiez2020implicit}]
The joint strategy $(h^\SE, m^\SE)\in\mc H \times \mc M$
is a \emph{differential Stackelberg equilibrium} if $m^S=\BR_M(h^S)$, the AI's best response given the Human's input $h^S$, and
\begin{align}
\frac{\partial c_H}{\partial h}(h^\SE,\BR_M(h^\SE))&=0,\ 
&\frac{\partial c_M}{\partial m}(h^\SE,m^\SE)=0,\\
\frac{\partial^2 c_H}{\partial h^2}(h^\SE,\BR_M(h^\SE))&\succ 0,\ 
&\frac{\partial^2 c_M}{\partial m^2}(h^\SE,m^\SE)\succ 0.
\end{align}

The learning algorithms we formulate are such that the agents follow myopic update rules which take steps in the direction of steepest descent for the
respective optimization problems.

\begin{figure*}[t]
\centering
\begin{minipage}{0.47\linewidth}
\begin{algorithm}[H]
\caption{Experiments with human subjects}\label{alg:human-AI}
\begin{algorithmic}
\Require initial $(h_{(0,:)}, m_{(0,:)})$, parameter $\alpha\ge0$
\For{$t=0,1,\dots,T-1$}
\Switch {Experiment}
\Case{1}
\State $\text{display }c_H(h_{(t,:)}, m_{(t,:)})$
\EndCase
\Case{2}
\State $\text{display  }\{c_H(h, m_{(t,:)})\ \mid \ h\in B_\epsilon(h_{(t,:)})\}$
\EndCase
\EndSwitch
\vspace{1.8em}
\State $h_{(t+1,:)} = {\tt manual\_input(t)} $ 
\Comment{Human input}
\State $m_{(t+1,:)} = m_t-\alpha  \tfrac{\partial c_M}{\partial m}(h_{(t,:)}, m_{(t,:)})$
\Comment{AI  update}
\EndFor
\end{algorithmic}
\end{algorithm}
\end{minipage}
\hspace*{0.25in}
\begin{minipage}{0.47\linewidth}
\begin{algorithm}[H]
\caption{Simulation of experiments}\label{alg:AI}
\begin{algorithmic}
\Require initial $(h_{(0,:)},m_{(0,:)})$, parameters $K,\alpha,\eta,\sigma>0$
\For{$t=0,1,\dots,T-1$}
\State $u_{t,0} = v_{t,0}= m_{(t,:)}$
\State $\delta  \sim N(0, \sigma^2 I_h)$
\For{$k=0,1,\dots,K-1$}
\State $u_{t,k+1} = u_{t,k}-\alpha  \tfrac{\partial c_M}{\partial m}(h_{(t,:)}+\delta,u_{t,k})$
\State $v_{t,k+1} = v_{t,k}-\alpha  \tfrac{\partial c_M}{\partial m}(h_{(t,:)}-\delta,v_{t,k})$
\EndFor
\State $g_t =  (c_H(h_{(t,:)} + \delta, u_{t,K}) - c_H(h_{(t,:)} - \delta, v_{t,K})) / \sigma^2 $
\State $h_{(t+1,:)} = h_{(t,:)}-\eta  g_t \delta $ 
\Comment{Human update}
\State $m_{(t+1,:)} = m_{(t,:)}-\alpha  \tfrac{\partial c_M}{\partial m}(h_{(t,:)}, m_{(t,:)})$
\Comment{AI update}
\EndFor
\end{algorithmic}
\end{algorithm}
\end{minipage}
\end{figure*}

\section{Experimental Setup}

This section contains details about the two-player continuous game's cost model, the experiment design, the AI adaptive algorithm, and the two cost feedback displays. A link directing to a copy of our experiment's game code can be found in the Supplemental Document (Section C.5).

\subsection{Game and Cost Models}

A Human and AI agent play a continuous game $(c_H,c_M)$ where cost functions $c_H$ and $c_M$ map from Human actions $h\in \mc H$ and AI actions $m\in \mc M$ to a real number. The costs are parameterized as quadratic functions given by
\begin{equation}
    \begin{aligned}
        c_H(h,m)=
        &\frac{1}{2}h^\top A_H h 
        + h^\top B_H m 
        + \frac{1}{2}m^\top D_H m \\
        &+ h^\top a_H
        + m^\top b_H
    \end{aligned}
    \label{eq:human-cost}
\end{equation}
and
\begin{equation}
    \begin{aligned}
        c_M(h,m)=
        &\frac{1}{2}m^\top A_M m 
        + m^\top B_M h 
        + \frac{1}{2}h^\top D_M h \\
        &+ m^\top a_M
        + h^\top b_M
    \end{aligned}
    \label{eq:AI-cost}
\end{equation}
where $A_H,B_H,D_H,a_H,b_H$ are the Human's cost parameters and $A_M,B_M,D_M,a_M,b_M$ are the AI agent's cost parameters, based on~\cite{chasnov2023human}. The sizes of the matrices are $A_H,B_H,D_M\in \R ^{d_H\times d_M}$, $A_M,B_M,D_H\in \R^{d_M\times d_H}$, $a_H,b_M\in\R^{d_H}$, and $a_M,b_H\in\R^{d_M}$. The square matrices $A_H,A_M$ are positive definite to ensure the existence of a Nash equilibrium, and $A_H-B_H A_{M}^{-1} B_M$ is positive definite to ensure the existence of a Stackelberg equilibrium. 

The parameters for the games in our experiments were selected such that the human actions at the game-theoretic equilibria were distinct positions, $h^{\NE}=(-0.25,-0.25)$ and $h^{\SE}=(+0.25,+0.25)$ ($h^{\NE}=-0.25$ and $h^{\SE}=+0.25$ for the one-dimensional Human agent input game). The parameters selected also had the Human and AI optimum cost at joint action vectors located at distinctly different locations in the joint action vectors,  $(h,m)$, resulting in the cost functions of the Human and AI to neither be fully aligned nor opposed.

\subsection{Human Experiment Design}
In this section, we describe the design of the experiment from the human participant's perspective. 

All human participants were recruited from the crowdsourcing platform, Prolific~\cite{palan2018prolific}. Participants had no prior training and were selected from the standard sample on the Prolific platform, distributed according to US and UK census data.

In the Prolific platform, participants were directed to an external link to our study. Studies either had the cost feedback display (Experiment 1) or the cost landscape display (Experiment 2), as shown in Figure~\ref{fig:setup}, and had participants play either the 1x2, 2x1, or 2x2 version of the game ($d_H \times d_M$ where $d_H$ is the dimensions of Human actions and $d_M$ is the dimensions of AI actions). Each participant was given only one type of feedback display and one type of joint action vector dimensions for the entirety of their participation. 

At the start of the study, participants were presented with an introduction screen explaining the task description and user instructions. Within each trial, participants were provided feedback information depending on the experiment, as described in Section~\ref{Cost Feedback Display}~and~\ref{Cost Landscape Feedback Display}. Participants were only instructed with the task to move their cursor to “keep the circle as small as possible” (Experiment 1 cost feedback display) or "keep the color inside the black circle cursor as light (close to white) as possible" (Experiment 2 cost landscape display). This task can be thought as being similar to using a dial to tune a radio station, using sound feedback to minimize radio static. When one trial ended, participants self-selected when to begin the next trial, optionally taking breaks, as needed.

For all experiments, participants provided manual input via a computer mouse (horizontal and vertical position of cursor) to continuously determine the value of a 2-dimensional input action $h\in[-1,1]^2\subset\mathbb{R}^2$. For experiment versions where the human had only one-input, participants only provided horizontal manual input to determine the value of a one-dimensional input action $h\in\mathbb{R}$. For the one-dimensional case, vertical movement was neither used to calculate the human's action nor recorded in our data. Within the code, this cursor input $h_{(t,:)} = (h_{(t,1)} , h_{(t,2)})$, where $t$ denotes the time and $h_{(t,1)}$ and $h_{(t,2)}$ denotes the horizontal and vertical input for the human, was scaled such that $h_{(t,:)} = (-1,-1)$ corresponded to the bottom-left corner of the participant's game display and $h_{(t,:)} = (+1,+1)$ corresponded to the top-right corner of the display. 

To help prevent human participants from memorizing the location of game equilibria, at the beginning of each trial a variable $s_i$ was chosen uniformly at random from $\{-1, +1\}$ and the map $h_{(t,i)} \to s_ih_{(t,i)}$ was applied to the participant's manual input for the duration of the trial, where $i$ denotes the horizontal ($i=1$) or vertical ($i=2$) input. When the variable's value was $s_i = -1$, this had the effect of applying a ``mirror'' symmetry to the input. Our experiments had four trials per adaptation rate, where $(s_1,s_2) = (1,1)$, $(s_1,s_2) = (-1,1)$, $(s_1,s_2) = (1,-1)$, and $(s_1,s_2) = (-1,-1)$.

Data was collected at a fixed rate (60 samples per second for Experiment 1 and 24 samples per second for Experiment 2) for a fixed number of trials depending on the feedback display and Human agent's input dimension. Each trial had a duration of 25 seconds.

\subsection{AI Adaptation Rule}

In this section, we describe the AI's adaptation rule.

In all experiments, the AI adapted its input actions using gradient descent,
\begin{equation}
m_{(t+1,:)} = m_{(t,:)} - \alpha \frac{\partial c_M}{\partial m}(h_{(t,:)},m_{(t,:)}),
\end{equation}
with one of the five different choices of adaptation rate \[
\alpha \in \{ 0, 0.001, 0.01, 0.1, 1\}.\]
At the slowest adaptation rate $\alpha = 0$, the AI implemented a constant policy, $m_{(t+1,:)} = m^{\NE}$, which was the AI's action component of the game's Nash equilibrium. At adaptation rates $\alpha>0$, the AI's input actions were initialized at a point away from both equilibria, $m_{(0,:)} = (0.1,0.1)$ ($m_{(0,:)} = (0.1)$ for the 2x1 version of the game). At the fastest adaption rate $\alpha = 1$, the AI implements the AI's best response to the human's action. Each adaptation rate condition was experienced once per symmetry (described in the previous section), in randomized order.

For the quadratic cost model ~\eqref{eq:AI-cost}, the AI's updates are linear in the players' actions. Given Human actions $h\in \R^{d_H}$, AI actions $m\in\R^{d_M}$, and AI cost parameters $A_M\in\R^{d_H\times d_H}, B_M\in\R^{d_M\times d_H}, a_M\in\R^{d_H}, b_M$, the AI adaptation rule is 
\begin{align}
m^+ &= \left\{\begin{array}{lr}
        m^{\NE}
        & \text{if } \alpha = 0, \\
        m - \alpha (A_M m + B_M h + a_M) 
        & \text{if } 0 < \alpha < 1, \\
        - A_M^{-1} (B_M h + a_M) 
        & \text{if } \alpha = 1.\\
        \end{array}\right.
        \label{eq:machineadaptation}
\end{align}
At each time step of the repeated interaction, the AI implemented a linear adaptation rule to update its action. The adaptation rule allowed the AI to adapt to the human's action in real-time, and defined the learning dynamics of the system. When the human played an action, the AI would respond with a new action. How the human responded to a change in the AI's action depended on the adaptation rate of the AI and what information was displayed to the human.

\subsection{Cost Feedback Display}\label{Cost Feedback Display}

For Experiment 1, participants were given feedback on their current cost, via an expanding/shrinking circle. The radial length of the circle represented the participant's cost at the current human-AI joint action vector, $c_{H}(h_{(t,:)},m_{(t,:)})$. Before and throughout each trial, participants were instructed to "keep this circle as small as possible". 

The cost feedback display was performed with the 1x2, 2x1, and 2x2 version of the game. Each version was tested with $n=30$ participants.

\subsection{Cost Landscape Feedback Display}\label{Cost Landscape Feedback Display}

For Experiment 2, participants were given feedback information on their localized cost landscape via a $7 \times 7$ grid of shaded dots centered at the cursor. The shaded dots represented the Human cost at the human-AI joint action, $\{ c_H(h, m_{(t,:)})\ \mid \ h\in B_\epsilon(h_{(t,:)})\}$, where the AI's action, $m_{(t,:)}$, was the AI's current action, and $B_\epsilon(h)=\{ h' \mid |h-h'|\leq \epsilon \}$ is box with side length $2\epsilon$ centered at the participant's action $h$. 

For the cost landscape feedback display, the lighter (closer to white) color of the dot represented positions of lower cost. Each colored dot represent the real-time position and Human cost for the participant's actions between $\pm 0.15$ of the participant's cursor, equally spaced by action values of $0.05$. Participants were instructed to "keep the color inside the black circle cursor as light (close to white) as possible".

The cost landscape feedback display was tested with $n=30$ participants on only the 2x2 version of the game.

\begin{figure}[t]
    \centering
    \subfloat[][$H$'s actions]{\includegraphics[width=0.33\linewidth]{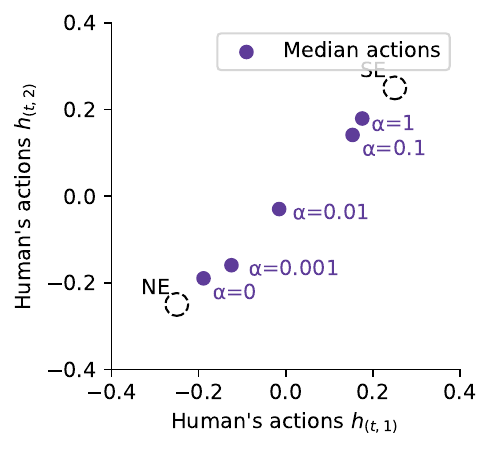}}
    \subfloat[][$M$'s actions]{\includegraphics[width=0.33\linewidth]{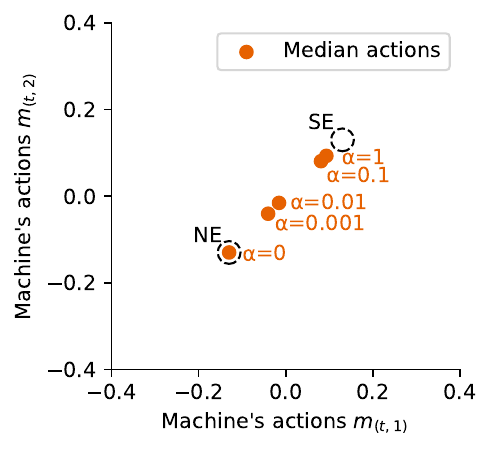}}
    \subfloat[][$H$ \& $M$'s costs]{\includegraphics[width=0.33\linewidth]{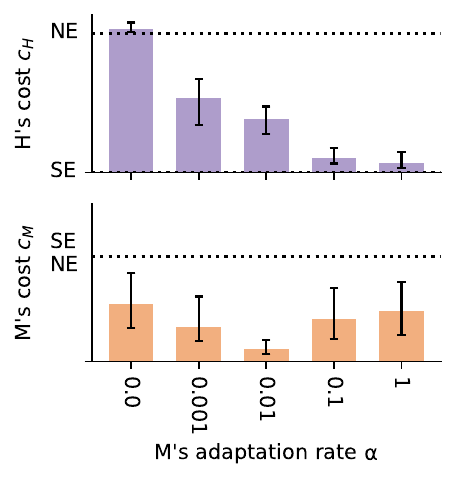}}
    
    \subfloat[][Histogram plots of $H$ \& $M$'s actions]{\includegraphics[width=0.99\linewidth]{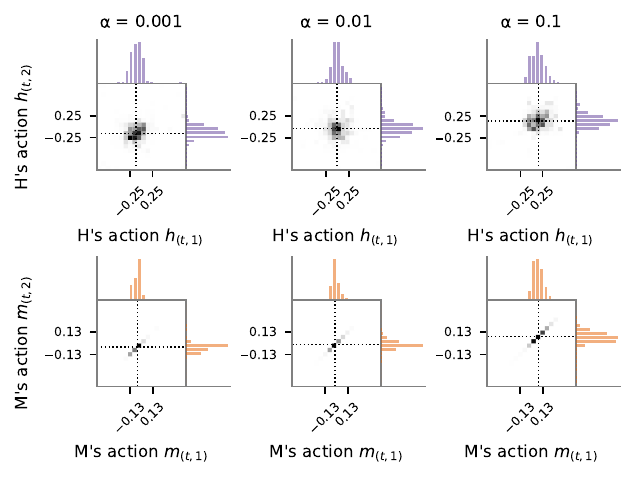}}
    \caption{Results of Experiment 1 (cost feedback) with $30$ participants.  
    The median actions and costs are compared to the Nash and Stackelberg equilibria. Experiment 1 shows increasing the AI's adaptation rates shifts the outcome from the Nash to the Stackelberg equilibrium.}
    \label{fig:exp1_2x2}
\end{figure}

\begin{figure}[t]
    \centering
    \subfloat[][$H$'s actions]{\includegraphics[width=0.33\linewidth]{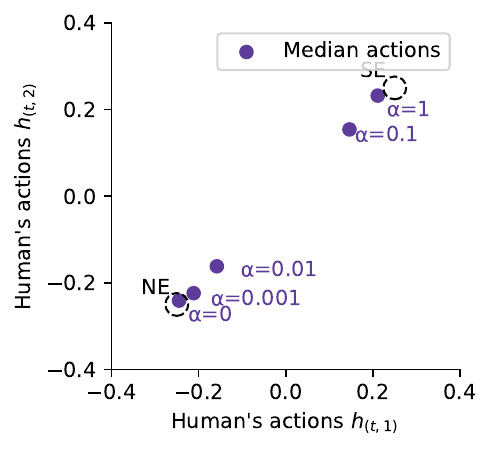}}
    \subfloat[][$M$'s actions]{\includegraphics[width=0.33\linewidth]{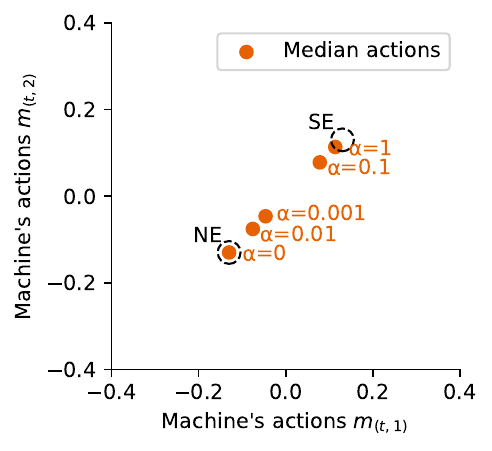}}
    \subfloat[][$H$ \& $M$'s costs]{\includegraphics[width=0.33\linewidth]{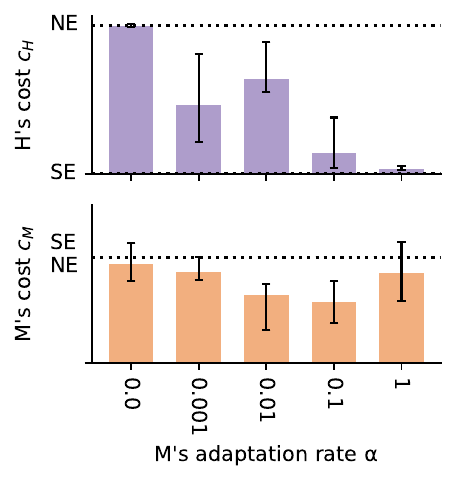}}

    \subfloat[][Histogram plots of $H$ \& $M$'s actions]{\includegraphics[width=0.99\linewidth]{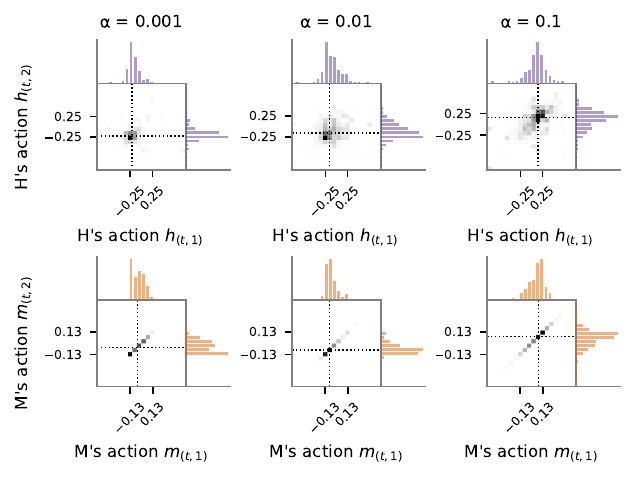}}
    \caption{Results of Experiment 2 (cost landscape feedback) with 30 participants. The median actions and costs are compared to the Nash and Stackelberg equilibria. Experiment 2, when compared to Experiment 1, shows a further shift towards the Nash equilibrium at slower adaptation rates.}
    \label{fig:exp2_2x2}
\end{figure}

\section{Results of Human Experiments}

We conducted two experiments with different populations of naive human subjects recruited from a crowdsourcing research platform. The participants engaged in a two-player human-AI game defined by a pair of quadratic cost functions $c_H$ and $c_M$ given in~\eqref{eq:human-cost} and \eqref{eq:AI-cost}. The AI adapted its actions within trials using gradient descent in~\eqref{eq:machineadaptation} while human participants were tasked with keeping their cost as small as possible using one of our two feedback displays. The experiment parameters were designed to yield distinct game-theoretic equilibria in the players' action spaces. These analytically determined equilibria were compared with the empirical distributions of actions reached by human participants and AI over a sequence of trials in each experiment. 

We tested five adaptation rates $\alpha \ge 0$ for the AI gradient descent algorithm, with multiple repetitions for each rate and the sequence of rates occurring in random order. In our experiments, we found that the observed behavior outcomes follow the predicted game-theoretic values. The following sections will focus on results from the 2x2 version of the experiment but similar behavior can be seen in the 1x2 and 2x1 versions of the experiment. We report figures demonstrating results of the 1x2 and 2x1 versions of the human participant experiment in the Supplemental Document (Section B). A link to our figure generation code and data collected can be found in the Supplemental Document (Section C.5).

\begin{figure}[hbt!]
    \center
    \includegraphics[width=0.9\linewidth]{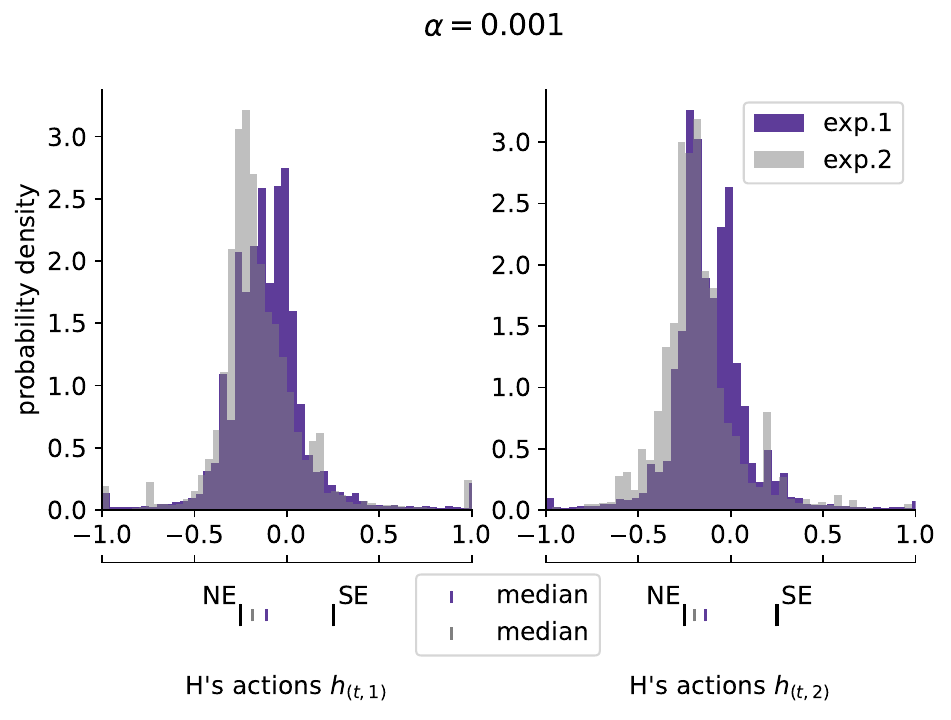}

    \includegraphics[width=0.9\linewidth]{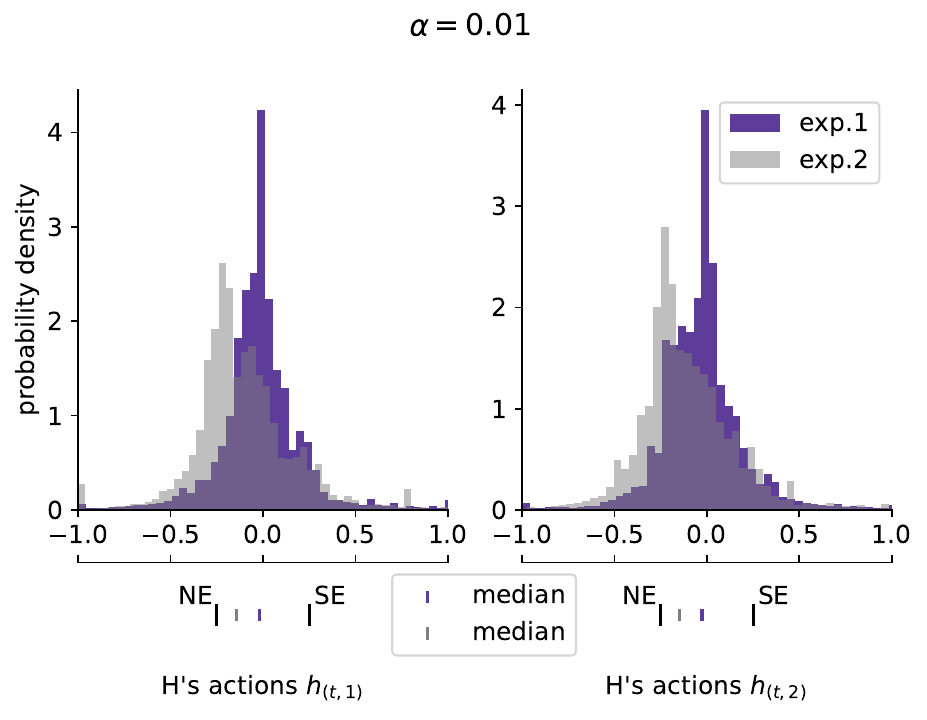}

    \includegraphics[width=0.9\linewidth]{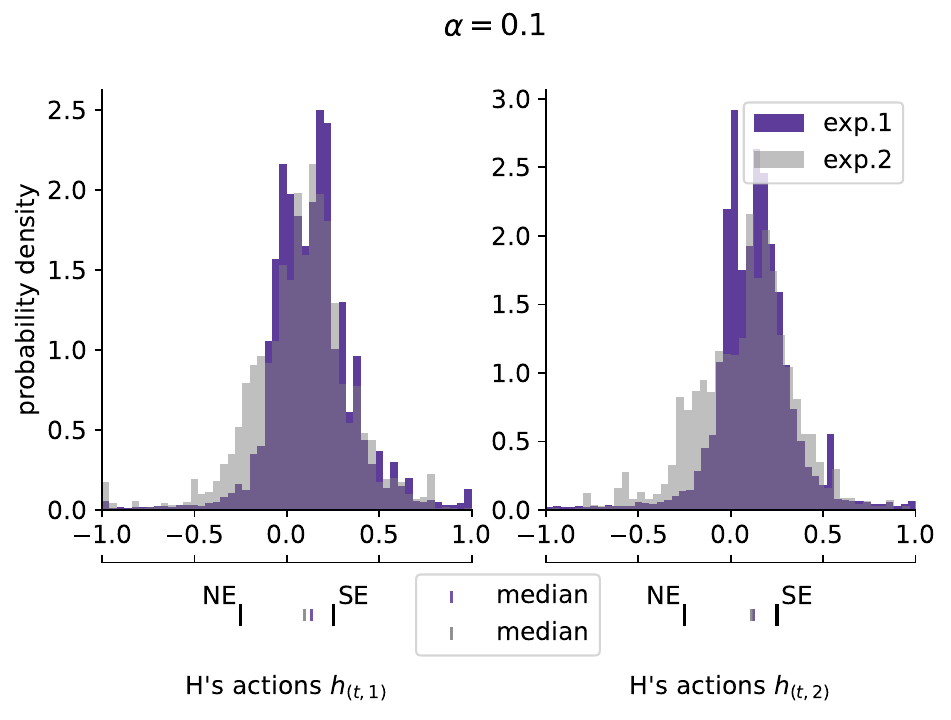}
    \caption{Comparing last 5 seconds of $H$'s actions with ticks displaying analytically calculated equilibria and median actions (Experiment 1 vs Experiment 2). Experiment 2 shows a shifted distribution of actions towards the Nash equilibrium.}
    \label{fig:exp1v2}
\end{figure}

\subsection{Cost Feedback Results}
\label{Cost Feedback Results}

In Experiment 1, participants were provided feedback on their current cost via an expanding/shrinking circle. We found that distributions of median $h$ and $m$ action vectors of the last 5 seconds of each trial ($n = 30$) shifted from close to the \textit{Nash equilibrium} (NE) at the slowest adaptation rate to close to the \textit{human-led Stackelberg equilibrium} (SE) at the fastest adaptation rate (Figure~\ref{fig:exp1_2x2}(a,b)). For the AI adaptation rates that we selected, median $h$ and $m$ action vectors were along a line between the analytically calculate NE and SE for each player: human action behavior near the NE for $\alpha = 0.001$, human action behavior near input $[0,0]$ for $\alpha = 0.01$, and human action behavior near the SE for $\alpha = 0.1$. The shift we observed from Nash to Stackelberg was in favor of the human resulting in a decrease in human cost as the joint human-AI action vectors shifted from NE to SE (Figure~\ref{fig:exp1_2x2}(c)). The AI's costs displayed a U-shape trend from slower adaptation rates to faster adaptation rates, resulting in a AI cost close to 0 (the AI agent's analytically calculated optimum cost) at $\alpha = 0.01$.

\subsection{Cost Landscape Feedback Results}
\label{Cost Landscape Feedback Results}

In Experiment 2, participants were provide information via a localized heat map of their cost landscape. This cost landscape display provided participants with additional information on the action dynamics of the game that was not available to participants in the cost feedback display. We found that distributions of median $h$ action vectors of the last 5 seconds of each trial ($n = 30$) resulted in a similar shift as Experiment 1: from near the NE at the slowest adaptation rate to near the SE at the fastest adaptation rate. Compared to the Experiment 1, the cost landscape experiment resulted in a general shift of median action vectors for the human participants towards the NE (Figure~\ref{fig:exp2_2x2}(a)). In Figure~\ref{fig:exp2_2x2}(c), human cost decrease as the adaptation rate increased, with an increase of human cost at $\alpha = 0.01$. For the AI, median $m$ action vectors displayed a similar behavior as the cost feedback experiment: with a shift from near NE at slow adaptation rates to near SE at fast adaptation rates with the exception of the adaptation rate of $\alpha = 0.001$, which was closer to SE than the faster adaptation rate of $\alpha = 0.01$ (Figure~\ref{fig:exp2_2x2}(b)). The AI's costs in the Experiment 2 displayed a similar U-shape outcome as the cost feedback experiment. Comparing the two feedback displays, the AI's costs in Experiment 2 are larger and closer to the analytically calculated NE and SE AI cost for all adaptation rates.

\section{Numerical Simulations}

The results found in the experiments (Section~\ref{Cost Feedback Results} and~\ref{Cost Landscape Feedback Results}) would not have been obtained if the Human agent also adapting its actions using gradient decent because changing adaptation rate alone in simultaneous gradient descent play does not change stationary points~\cite{pmlr-v115-chasnov20a}. We found that simulating the human adaptation as a two-point zeroth-order approximation of the Human agent's gradient provided similar learning dynamics as our human experiments results. 

Along with numerical simulations for the 1x2, 2x1, and 2x2 versions of our human experiments, we implemented numerical simulations to demonstrate the learning dynamics on a larger scale problem. The dimensions of players' action spaces are $d_H=64$ and $d_M=128$. The pseudocode of the simulation is listed in Algorithm~\ref{alg:AI}, which calculated a two-point zeroth-order approximation of the Human agent's gradient~\cite{nesterov2017random}. This section contains details about the 64x128 numerical simulations and results. A link to all numerical simulation code can be found in the Supplemental Document (Section C.5).

\subsection{Varying Adaptation Rate in Simulation}
The 64x128 numerical simulations tested three AI adaptation rates --- slow ($\alpha = 0.0001$), moderate ($\alpha = 0.001$), and fast ($\alpha = 0.01$) --- while the simulated Human agent had adaptation rate $\eta=0.01$. In the simulations, the number of iterations was $T=1000$, representing to length of a trial in the experiment while the inner loop $K$ was set to $K=10$ to simulation the Human's strategy to test an action near their current action in order to gain cost information to estimate which action to take next. At each trial, the simulated Human selects a random variable $\delta$ from a normal distribution with standard deviation $\sigma$. The direction $\delta$ is used in the model of the Human's gradient update. The cost parameters for the simulation were selected at random, following the same criteria as the human experiments.

\subsection{Results of Numerical Simulation} 

\begin{figure}[t]
    \center
    \subfloat[][distance of $H$'s actions from equilibria for  adaptation rates $\alpha$]{\includegraphics[width=0.99\linewidth]{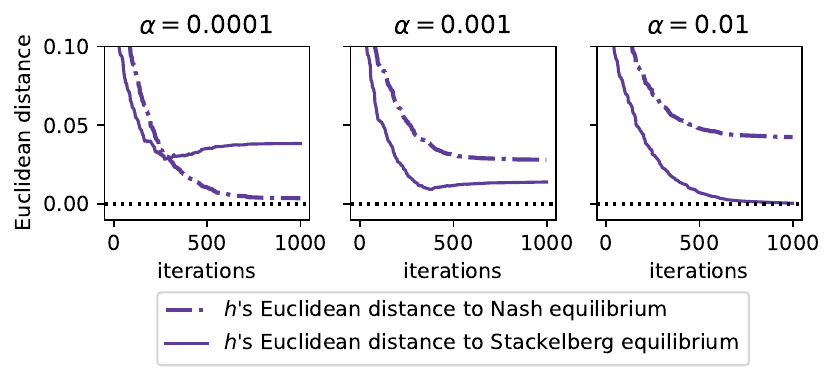}}
    
    \subfloat[][distance of $M$'s actions from  equilibria for  adaptation rates $\alpha$]{\includegraphics[width=0.99\linewidth]{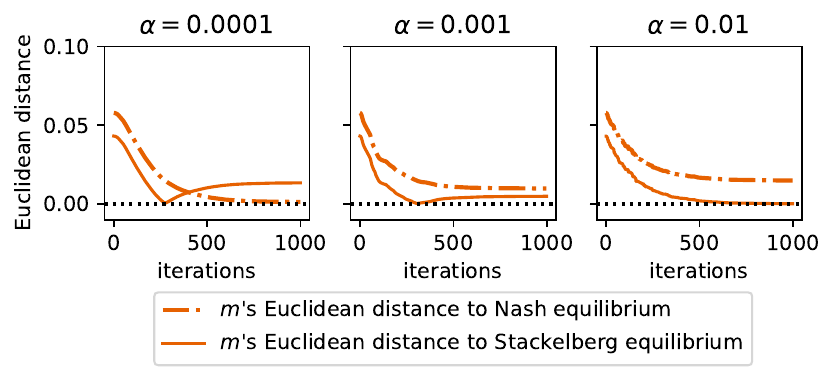}}
    \caption{Results of simulating the learning dynamics of a game with action vector dimensions of 64x128. For slow adaptation rates, the actions shift towards the Nash equilibrium, while for fast adaptation rates, the actions shift towards the Stackelberg equilibrium, matching the experiments.  
    }
    \label{fig:simresults}
\end{figure}

Figure~\ref{fig:simresults} shows results from our numerical simulation for the 64 x 128 game with $\alpha \in \{0.0001,0.001,0.01\}$. Without changing the adaptation rules defined in Algorithm~\ref{alg:AI}, both the simulated Human and AI converged to the NE equilibrium at the slowest adaption rate and the SE equilibrium at the fastest adaptation rate. The simulation matches the same trend we witnessed in Experiment 1 and 2: shift from NE at the slowest adaptation rate to the SE at the fastest adaptation rate.

\section{Discussion}

Our study focused on the effect of elements on both sides of human-AI interactions: adaption rate (the AI side) and user-feedback information (the human side).

In Experiment 1, we observed a correlation between adaptation rates and human cost: as the AI's adaptation rate increased, the human's cost decreased (Figure~\ref{fig:exp1_2x2}). These findings in the multi-dimensional game are consistent with that of the one-dimensional game in ~\cite{chasnov2023human}. This suggests that quicker AI responses allow humans to better anticipate AI actions, improving human performance.

Experiment 2 kept all elements of the game the same as Experiment 1 but explored the effects of providing the participant additional information which could be used to estimate the dynamics of their cost landscape. We observed a similar correlation between adaptation rate and human cost at the slow ($\alpha = 0.001$) and fast ($\alpha = 0.1$) adaptation rates but observed an increase in human cost at the moderate ($\alpha = 0.01$) adaptation rate (Figure~\ref{fig:exp2_2x2}). Looking at the human's actions alone, median human action outcomes still follow the pattern shown in Experiment 1, shifting from NE at slower AI adaptation rates to SE at faster AI adaptation rates. Additionally, Experiment 2 showed a general shift of human action outcomes towards the Nash equilibrium in Experiment 2. This may be because the cost landscape feedback suggest to participants that $h$ actions towards the NE will result in lower cost with lighter (closer to white) dots in the direction of the NE and darker dots in the direction away from NE, these results suggests a change in the human adaptation strategy between focusing on the cost landscape information or the current cost information (color of the cursor). 

During slower adaptation rates, participants may not be able to anticipate the AI's responses (Experiment 1), resulting in more trust in the heat map information. This is because the heat map information displays the localized Human cost landscape at the human's slice of the cost landscape constrained to the AI's current action. This means when participants moves their actions in the direction of lighter color in the heat map, the cursor color will reflect an expected change due to the AI adapting slowly to the human's new input. This is different for the faster adaptation rates because as the participant moves their actions in the direction of light color, the AI changes its input quickly, which creates an unexpected change in cursor color. Because of this, participants playing with AIs at faster adaptation rates may choose to focus on their cursor color to achieve smaller cost.

Our experiments suggest that humans may give priority for cost landscape feedback information when they are less able to anticipate AI actions. Future work can investigate use biomarkers like eye tracking, which has been shown to be predictors of human intent~\cite{aronson2022gaze}, to determine what feedback information the participant is using and actively to change AI adaptation rate to shift human behavior towards the Nash or Stackelberg equilibrium.

The cost model and adaptation rule used in our experiments aims to be representative of real-world scenarios where agents dynamically make decisions in continuous spaces. However, our experiments are limited to the specific user feedback and two-dimensional manual input chosen. In real-world scenarios, the action spaces may be larger and costs may not be observable. Future work will address these limitations by studying larger models and studying other classes action spaces, such as using electromyography or eye-tracking devices as continuous inputs for the human, and decoders or controllers as the AI.

\section{Conclusion}

In order to prevent undesirable behavior and maximize benefit to individuals and society, it is critical to develop better predictors of human behavior in the interactions between humans and AI. Here we showed that a specific expansion to the information available to the human, from single point cost feedback to multi-point cost landscape, resulted in a shift of median actions across different AI adaptation rates towards the Nash equilibrium while still following the same trend of a shift from the Nash equilibrium at slower adaptation rates to the human-led Stackelberg equilibrium at faster adaptation rates. This work highlights the importance of validating behavioral models over different user feedback displays and adaptation parameters, opening the door to future research of human-AI co-adaptive interactions.

\newpage

\bibliographystyle{named}
\bibliography{main}

\newpage
\onecolumn
\appendix

\begin{center}
{\sf \Huge Supplemental Document for\\"Effect of Adaptation Rate and Cost Display in a Human-AI Interaction Game"}
\end{center}

\section{Cost Parameters}

In this section, we provide the parameter values for the cost models used in our experiments.

\subsection{2x2 experiment parameter values}

The parameter values are chosen to be $A_H=A_M=I$ are 2x2 identity matrices and
\begin{equation}
B_H = \begin{bmatrix}
  0.0677 & 0.0677 \\
  0.1182 & 0.1182 
  \end{bmatrix},
D_H = \begin{bmatrix}
  2.0006 & -0.9800 \\
  -0.9800 & 2.0644 
  \end{bmatrix},
a_H = \begin{bmatrix}
  0.2254 & 0.2254
  \end{bmatrix},
b_H = \begin{bmatrix}
  0.9556 & 0.9300
  \end{bmatrix},
\end{equation}
\begin{equation}
B_M = \begin{bmatrix}
  0.2650 & 0.2650 \\
  0.2654 & 0.2654 
  \end{bmatrix},
D_M = \begin{bmatrix}
  11.1467 & 6.4300 \\
  6.4300 & 5.7645 
  \end{bmatrix}
a_M = \begin{bmatrix}
  0 & 0
  \end{bmatrix},
b_M = \begin{bmatrix}
  0 & 0
  \end{bmatrix}
\end{equation}

\noindent These parameters were chosen such that the players' optima and the constellation of relevant game-theoretic equilibria were distinct positions:

\begin{align*}
\text{Human optimum } (h_H^{*},m_H^{*}) &= ( [-0.06, -0.06], 
[-0.90, -0.87] ),\\
\text{AI optimum } (h_M^{*},m_M^{*}) &= ( [0, 0], 
[0, 0] ),\\
\text{Nash equilibrium } (h^\NE,m^\NE) &= ( [-0.25, -0.25], 
[0.13, 0.13] ),\\
\text{Human-led Stackelberg equilibrium } (h^\SE,m^\SE) &= ( [+0.25,  +0.25],  
[-0.13,  -0.13])\\
 \end{align*}

\subsection{1x2 experiment parameter values}

The parameter values are chosen to be $A_H= I$, $A_M=I$; where $A_H$ is a 1x1 identity matrix and $A_M$ is a 2x2 identity matrix, and 

\begin{equation}
B_H = \begin{bmatrix}
  0.1889 \\
  -0.5000
  \end{bmatrix},
D_H = \begin{bmatrix}
  1.4145 & 0.2500 \\
  0.2500 & 1.2000 
  \end{bmatrix},
a_H = \begin{bmatrix}
  0.2973
  \end{bmatrix},
b_H = \begin{bmatrix}
  0.5290 & 1
  \end{bmatrix},
\end{equation}
\begin{equation}
B_M = \begin{bmatrix}
  0.6082 & 0.6082 
  \end{bmatrix},
D_M = \begin{bmatrix}
  4.0400
  \end{bmatrix}
a_M = \begin{bmatrix}
  0 & 0
  \end{bmatrix},
b_M = \begin{bmatrix}
  0
  \end{bmatrix}
\end{equation}

\noindent These parameters were chosen such that the players' optima and the constellation of relevant game-theoretic equilibria were distinct positions:

\begin{align*}
\text{Human optimum } (h_H^{*},m_H^{*}) &= ( [-0.89], 
[-0.04, -1.19] ),\\
\text{AI optimum } (h_M^{*},m_M^{*}) &= ( [0], 
[0, 0] ),\\
\text{Nash equilibrium } (h^\NE,m^\NE) &= ( [-0.25], 
[0.15, 0.15] ),\\
\text{Human-led Stackelberg equilibrium } (h^\SE,m^\SE) &= ( [+0.25],  
[-0.15,  -0.15])\\
\end{align*}

\subsection{2x1 experiment parameter values}

The parameter values are chosen to be $A_H= I$, $A_M=I$; where $A_H$ is a 2x2 identity matrix and $A_M$ is a 1x1 identity matrix, and

\begin{equation}
B_H = \begin{bmatrix}
  -0.3200 & -0.3200 
  \end{bmatrix},
D_H = \begin{bmatrix}
  1.2274 
  \end{bmatrix},
a_H = \begin{bmatrix}
  0.3677 & 0.3677
  \end{bmatrix},
b_H = \begin{bmatrix}
  1.611
  \end{bmatrix},
\end{equation}
\begin{equation}
B_M = \begin{bmatrix}
  0.7358 \\
  0.7358
  \end{bmatrix},
D_M = \begin{bmatrix}
  15.8090 & -0.2828 \\
  -0.2828 & 1.0000 
  \end{bmatrix}
a_M = \begin{bmatrix}
  0
  \end{bmatrix},
b_M = \begin{bmatrix}
  0 & 0
  \end{bmatrix}
\end{equation}

\noindent These parameters were chosen such that the players' optima and the constellation of relevant game-theoretic equilibria were distinct positions:

\begin{align*}
\text{Human optimum } (h_H^{*},m_H^{*}) &= ( [-0.95, -0.95], 
[-1.81] ),\\
\text{AI optimum } (h_M^{*},m_M^{*}) &= ( [0, 0], 
[0] ),\\
\text{Nash equilibrium } (h^\NE,m^\NE) &= ( [-0.25, -0.25], 
[0.37] ),\\
\text{Human-led Stackelberg equilibrium } (h^\SE,m^\SE) &= ( [+0.25,  +0.25],  
[-0.37])\\
\end{align*}

\section{Additional Results}

Below are results pertaining to the 1x2 and 2x1 versions of the human-AI experiment.

\begin{figure}[hbt!]
  \centering
  \subfloat[][]{\includegraphics[width=0.24\linewidth]{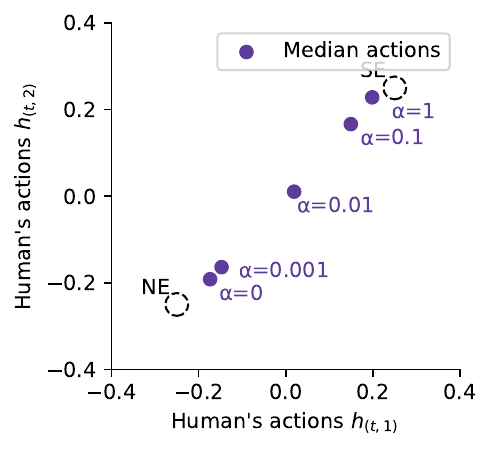}}
  \subfloat[][]{\includegraphics[width=0.24\linewidth]{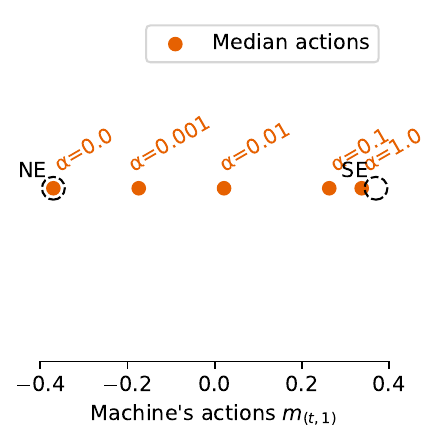}}
  \subfloat[][]{\includegraphics[width=0.24\linewidth]{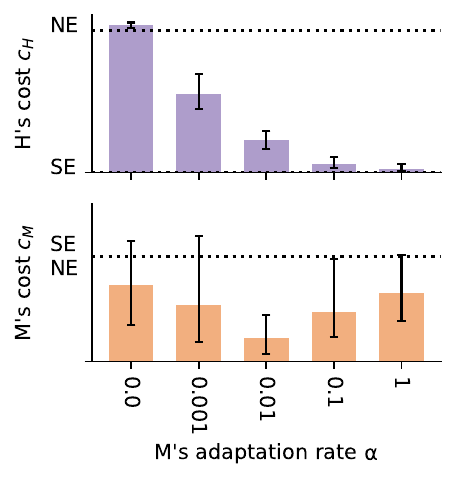}}

  \subfloat[][]{\includegraphics[width=0.72\linewidth]{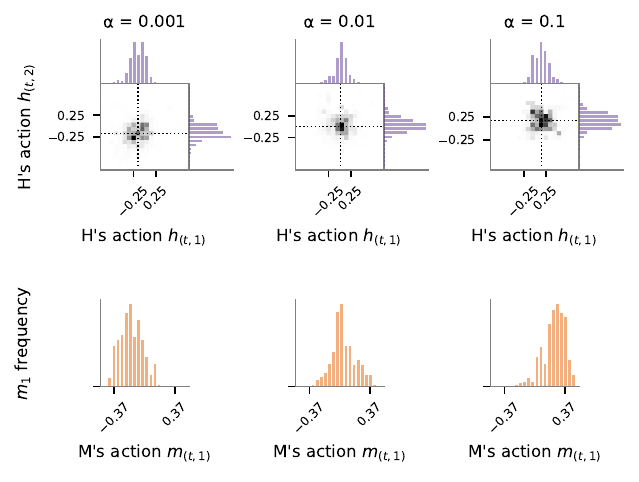}}
  \caption{Human cost feedback (Experiment 1, $n = 30$). (\textbf{a}) Median human actions for each AI adaptation rate overlaid on game-theoretic equilibria. (\textbf{b}) Median AI actions for each AI adaptation rate overlaid on game-theoretic equilibria. (\textbf{c}) Cost distributions for each AI adaptation rate displayed using box plots with error bars showing 25th, 50th, and 75th percentiles. (\textbf{d}) One- and two-dimensional histograms of actions for different adaptation rates with game-theoretic equilibria overlaid.}
  \label{fig:exp1_2x1}
\end{figure}

\begin{figure}[hbt!]
  \centering
  \subfloat[][]{\includegraphics[width=0.24\linewidth]{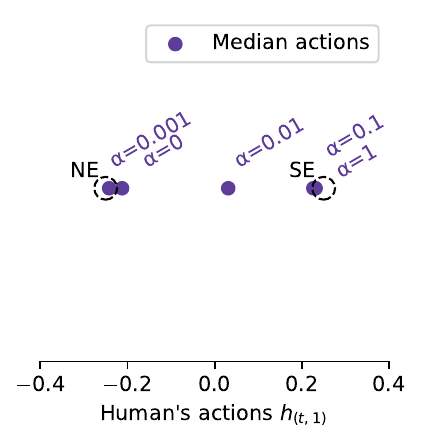}}
  \subfloat[][]{\includegraphics[width=0.24\linewidth]{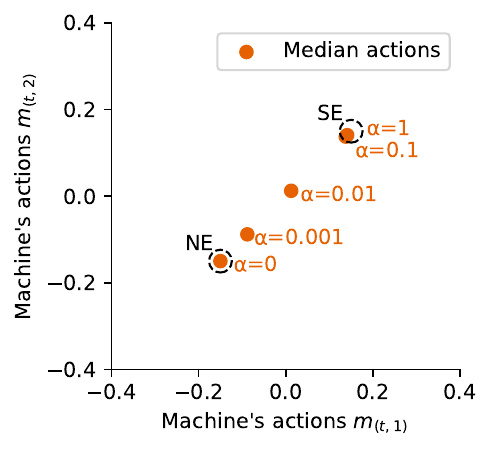}}
  \subfloat[][]{\includegraphics[width=0.24\linewidth]{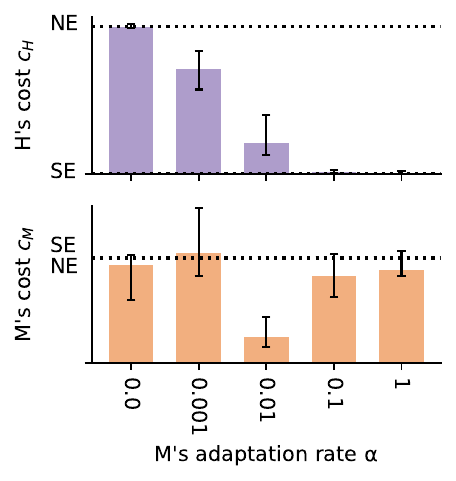}}

  \subfloat[][]{\includegraphics[width=0.72\linewidth]{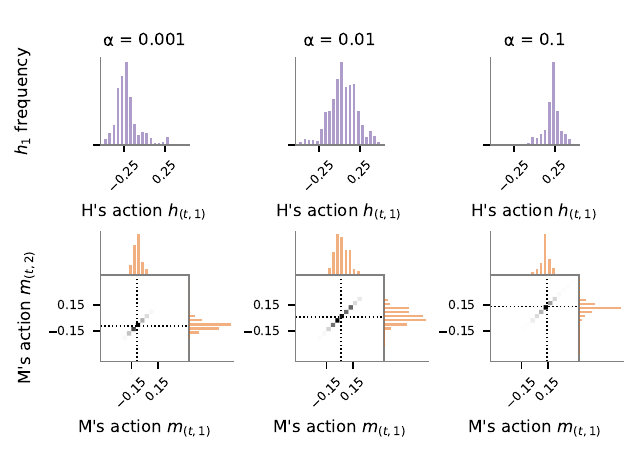}}
  \caption{Human cost feedback (Experiment 1, $n = 30$). (\textbf{a}) Median human actions for each AI adaptation rate overlaid on game-theoretic equilibria. (\textbf{b}) Median AI actions for each AI adaptation rate overlaid on game-theoretic equilibria. (\textbf{c}) Cost distributions for each AI adaptation rate displayed using box plots with error bars showing 25th, 50th, and 75th percentiles. (\textbf{d}) One- and two-dimensional histograms of actions for different adaptation rates with game-theoretic equilibria overlaid.}
  \label{fig:exp1_1x2}
\end{figure}

\newpage

\section{Additional Experimental Details}

This section includes additional information about experiment details not already included the the main paper.

\subsection{Prolific.com}

All human participant experiments were conducted on the Prolific.com crowd-sourced research website. The participants were recruited and paid, an estimated 12.00 USD per hour for a 10-15 minute study, on the Prolific.com website. On this site, we selected to recruit participants from a standard sample (distributed sample of available participants) in the United States and United Kingdom, according to census data. Participants were only allowed to participate a single time in one of our experiments while also requiring that the study was completed on a desktop device.

\subsection{Web Browser Client}

All experiments were written in JavaScript, HTML, and CSS. We have an HTML file for each version of the experiment (each combination of 1x2, 2x1, and 2x2 dimensions). We test five adaptation rates, $\alpha = 0, 0.001, 0.01, 0.1, 1$, and for each adaptation rate we will flip the horizontal and vertical axis based on the setting of the game. For example, game setting 1x2 means the calculation involves the human's actions in the horizontal direction only and the AI's actions in both horizontal and vertical direction. Game version 1x2 means we will give another set of trials for the flipped horizontal human input. Thus, the number of trials for game setting 1x2 is 10 (5 learning rates times 2 for the original and flipped horizontal input). Thus, game setting 2x1 and 2x2 will flip the human's actions in horizontal and vertical directions and have 20 total trials.

A loop in the embedded JavaScript code handled every trial as an iteration. At each time step of a trial, the program collected the human participant's cursor position as the Human's input, along with the AI's input, the Human's cost, the AI's cost, the AI's adaptation rate, and the time step. Data from each single trial were collected as an array of values. After all trials for one participant were finished, the arrays were converted to a single JSON object and sent to the database.

\subsection{Data Collection Server}

The URL distributed to experiment participants were hosted for free on Replit. Specifically, an API endpoint was established using FastAPI to continuously listen for participant connections. During a participant's session, a JSON object is sent back to the Replit server following each trial. A Sqlite3 database was set up in Replit to store JSON data from participants. To distinguish data from different participants, each participant was assigned a unique key, provided by the Prolific website, at the beginning of the experiment session that is later associated with every JSON object uploaded to the database.

\subsection{Data Processing Scripts}

The experimental data were initially gathered in JSON format and subsequently transformed into table format and stored as CSV files by the data processing scripts for visualization. The CSV files were structured such that rows corresponded to the trials, while columns represented the data attributes, such as pubkey, $\text{time step } t$, $h_{(t,1)}$, $h_{(t,2)}$, $m_{(t,1)}$, $m_{(t,2)}$, and so forth. 

Data collected in CSV files were processed and visualized using Jupyter Notebooks. CSV data were parsed into DataFrame object of the same format. All trials were trimmed to preserve only the data from the last five seconds. For each version of the game, the data was used to produce a bar plot displaying human's cost and AI's cost against the AI's adaptation rates, a dot plot displaying the human's median actions for each adaptation rate, a dot plot displaying the AI's median actions for each adaptation rate, and a collection of histograms, each displaying the human's and AI's actions from all trials corresponding to an adaptation rate.

\subsection{Code and Data}
\label{Code and Data}

Game code for all the experiments can be found in the following link to a GitHub repository, \url{https://anonymous.4open.science/r/Effect-of-Adaptation-Rate-and-Cost-Display-in-a-Human-AI-Interaction-Game-E855/README.md}. The README.md file has information on using the game code files.

Analysis code and simulation code for all the experiments can be found in the following link, \url{https://anonymous.4open.science/r/Effect-of-Adaptation-Rate-and-Cost-Display-in-a-Human-AI-Interaction-Game-E855/README.md}. (This link is currently missing data CSV files for all of the experiments. This will be included when this code is moved to a CodeOcean reproducible capsule, following the double-blind review process.)

\end{document}